\documentclass[10pt,twocolumn,letterpaper]{article}

\usepackage{cvpr}
\usepackage{times}
\usepackage{epsfig}
\usepackage{graphicx}
\usepackage{amsmath}
\usepackage{amssymb}
\usepackage{relsize}
\usepackage{caption,subcaption}
\usepackage{verbatim}

\newcommand{\algname}{{VoxelMorph}} 
\DeclareMathOperator*{\argmin}{arg\,min}

\usepackage[pagebackref=true,breaklinks=true,letterpaper=true,colorlinks,bookmarks=false]{hyperref}

\cvprfinalcopy 

\def\cvprPaperID{3262} 

\ifcvprfinal\fi
\begin{document}

\title{An Unsupervised Learning Model for \\Deformable Medical Image Registration}

\author{Guha Balakrishnan\\
MIT\\
{\tt\small balakg@mit.edu}\\
\and
Amy Zhao\\
MIT\\
{\tt\small xamyzhao@mit.edu}
\and
Mert R. Sabuncu\\
Cornell University\\
{\tt\small msabuncu@cornell.edu}
\and
John Guttag\\
MIT\\
{\tt\small guttag@mit.edu}
\and
Adrian V. Dalca\\
MIT and MGH\\
{\tt\small adalca@mit.edu}\\
}

\maketitle

\begin{abstract}
We present a fast learning-based algorithm for deformable, pairwise 3D medical image registration. Current registration methods optimize an objective function independently for each pair of images, which can be time-consuming for large data. We define registration as a parametric function, and optimize its parameters given a set of images from a collection of interest. Given a new pair of scans, we can quickly compute a registration field by directly evaluating the function using the learned parameters. We model this function using a convolutional neural network (CNN), and use a spatial transform layer to reconstruct one image from another while imposing smoothness constraints on the registration field. The proposed method does not require supervised information such as ground truth registration fields or anatomical landmarks. We demonstrate registration accuracy comparable to state-of-the-art 3D image registration, while operating orders of magnitude faster in practice. Our method promises to significantly speed up medical image analysis and processing pipelines, while facilitating novel directions in learning-based registration and its applications. Our code is available at https://github.com/balakg/voxelmorph.
\end{abstract}

\section{Introduction}
Deformable registration is a fundamental task in a variety of medical imaging studies, and has been a topic of active research for decades. In deformable registration, a dense, non-linear correspondence is established between a pair of $n$-D image volumes, such as 3D MR brain scans, depicting similar structures. Most registration methods solve an optimization problem for each volume pair that aligns voxels with similar appearance while enforcing smoothness constraints on the registration mapping. Solving this optimization is computationally intensive, and therefore extremely slow in practice. 

In contrast, we propose a novel registration method that learns a parametrized registration \emph{function} from a collection of volumes. We implement the function using a convolutional neural network (CNN), that takes two $n$-D input volumes and outputs a mapping of all voxels of one volume to another volume. The parameters of the network, i.e., the convolutional kernel weights, are optimized using a training set of volume pairs from the dataset of interest. By sharing the same parameters for a collection of volumes, the procedure learns a common representation which can align any new pair of volumes from the same distribution. In essence, we replace a costly optimization of traditional registration algorithms for each test image pair with one global function optimization during a training phase. Registration between a new test scan pair is achieved by simply evaluating the learned function on the given volumes, resulting in rapid registration.

The novelty of this work is that:

\begin{itemize}
\item we present a learning-based solution requiring no supervised information such as ground truth correspondences or anatomical landmarks during training,
\item we propose a CNN function with parameters shared across a population, enabling registration to be achieved through a function evaluation, and
\item our method enables parameter optimization for a variety of cost functions, which can be adapted to various tasks.
\end{itemize}

Throughout this paper, we use the example of registering 3D MR brain scans. However, our method is broadly applicable to registration tasks, both within and beyond the medical imaging domain. We evaluate our method on a multi-study dataset of over 7,000 scans containing images of healthy and diseased brains from a variety of age groups. Results show that our method achieves comparable accuracy to a state-of-the-art registration package, while taking orders of magnitude less time. Scans that used to take two hours to register can now be registered within one or two minutes using a CPU, and under a second with a GPU. This is of significant practical importance for many medical image analysis tasks. 
\section{Background}
\begin{figure}[h!]
\begin{center}
\includegraphics[width=\linewidth]{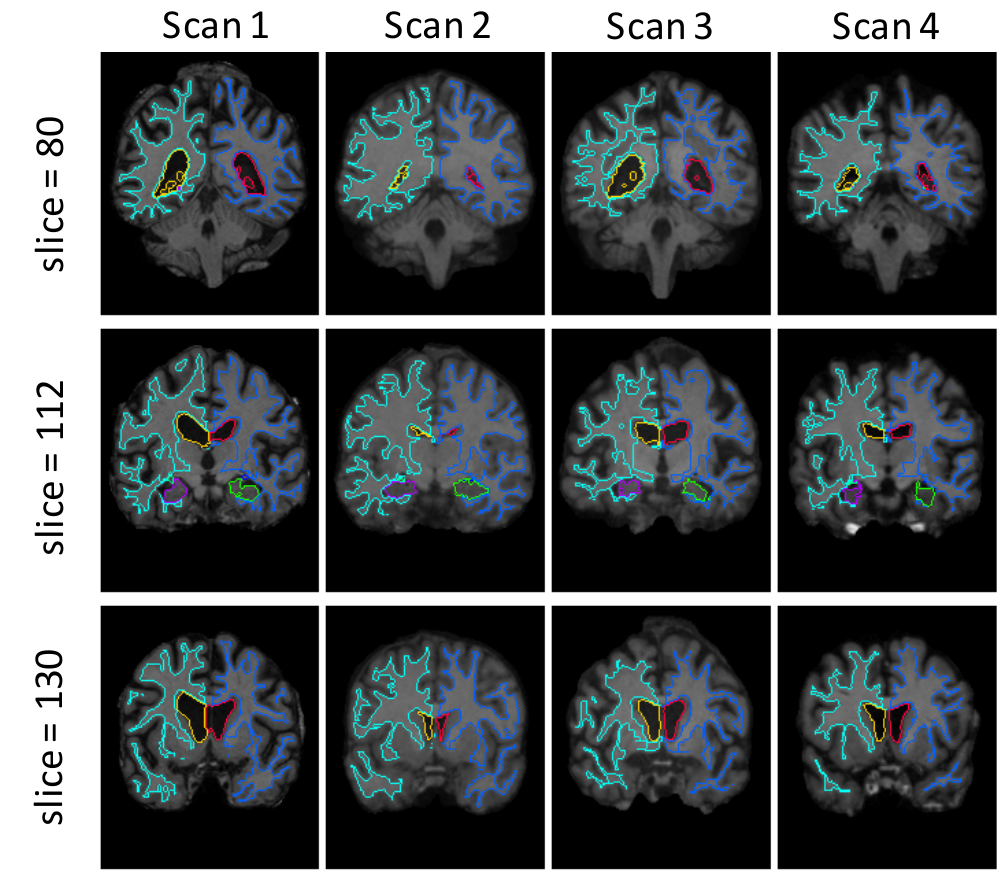}
\end{center}
\caption{Example coronal slices from the 3D MRI brain dataset, after affine alignment. Each column is a different scan (subject) and each row is a different coronal slice. Several significant anatomical regions are outlined using different colors: L/R white matter in light/dark blue, L/R ventricles in yellow/red, and L/R hippocampi in purple/green. There are significant structural differences across scans, necessitating a deformable registration step to analyze inter-scan variations.}
\label{fig:examples}
\end{figure}

In the typical volume registration formulation, one (moving or source) volume is warped to align with a second (fixed or target) volume. Deformable registration strategies separate an initial affine transformation for global alignment from a typically much slower deformable transformation with higher degrees of freedom. We concentrate on the latter step, in which we compute a dense, nonlinear correspondence for all voxels. Fig.~\ref{fig:examples} shows sample 2D coronal slices taken from 3D MRI volumes, with boundaries of several anatomical structures outlined. There is significant variability across subjects, caused by differences in health state and natural anatomical variations in healthy brains. Deformable registration enables comparison of structures across scans and population analyses. Such analyses are useful for understanding variability across populations or the evolution of brain anatomy over time for individuals with disease.

Most existing registration algorithms iteratively optimize a transformation based on an energy function. Let $F,M$ denote the fixed and moving images, respectively, and let $\phi$ be the registration field. The optimization problem is typically written as:
\begin{align}
\label{eqn:energy}
\hat{\phi} &= \argmin_{\phi}\mathcal{L}(F,M, \phi), 
\end{align} 
\noindent where 
\begin{align}
\label{eqn:loss}
\mathcal{L}(F,M, \phi) = \mathcal{L}_{sim}(F,M(\phi)) + \lambda \mathcal{L}_{smooth}(\phi), 
\end{align}

\noindent $M(\phi)$ is $M$ warped by $\phi$, function $\mathcal{L}_{sim}(\cdot,\cdot)$ measures image similarity between $M(\phi)$ and $F$, $\mathcal{L}_{smooth}(\cdot)$ imposes regularization on $\phi$, and $\lambda$ is the regularization parameter. 

There are several common formulations for $\phi$, $\mathcal{L}_{sim}$ and $\mathcal{L}_{smooth}$. Often, $\phi$ is a displacement vector field, specifying the vector offset from $F$ to $M$ for each voxel.  Diffeomorphic transforms are a popular alternative that model $\phi$ as the integral of a velocity vector field. As a result, they are able to preserve topology and enforce invertibility on $\phi$. Common metrics used for $\mathcal{L}_{sim}$ include mean squared voxel difference, mutual information, and cross-correlation. The latter two are particularly useful when volumes have varying intensity distributions and contrasts. $\mathcal{L}_{smooth}$ enforces a spatially smooth deformation, often modeled as a linear operator on spatial gradients of $\phi$. In our work, we optimize function parameters to minimize the expected energy of the form of~\eqref{eqn:energy} using a dataset of volume pairs, instead of doing it for each pair independently.

\section{Related Work}

\subsection{Medical Image Registration (Non-learning-based)}
There is extensive work in 3D medical image registration~\cite{ashburner2007,avants2008,bajcsy1989,beg2005,dalca2016,glocker2008,thirion1998}.\footnote{in medical imaging literature, the volumes produced by 3D imaging techniques are often referred to as images} Several studies optimize within the space of displacement vector fields. These include elastic-type models~\cite{bajcsy1989,shen2002}, statistical parametric mapping~\cite{ashburner2000}, free-form deformations with b-splines,~\cite{rueckert1999} and Demons~\cite{thirion1998}. Our model also assumes displacement vector fields. Diffeomorphic transforms, which are topology-preserving, have shown remarkable success in various computational anatomy studies. Popular formulations include Large Diffeomorphic Distance Metric Mapping (LDDMM)~\cite{beg2005}, DARTEL~\cite{ashburner2007} and standard symmetric normalization (SyN)~\cite{avants2008}.

\subsection{Medical Image Registration (Learning-based)}
There are several recent papers proposing neural networks to learn a function for medical image registration. Most of these rely on ground truth warp fields or segmentations~\cite{krebs2017,rohe2017,sokooti2017,yang2017}, a significant drawback compared to our method, which does not require either. Two recent works~\cite{devos2017,li2017} present unsupervised methods that are closer to our approach. Both propose a neural network consisting of a CNN and spatial transformation function~\cite{jaderberg2015} that warps images to one another. Unfortunately, these methods are preliminary and have significant drawbacks: they are only demonstrated on limited subsets of volumes, such as 3D subregions or 2D slices, and support only small transformations. Others~\cite{devos2017} employ regularization only implicitly determined by interpolation methods. In contrast, our generalizable method is applicable to entire 3D volumes, handles large deformations, and enables any differentiable cost function. We present a rigorous analysis of our method, and demonstrate results on full MR volumes.

\subsection{2D Image Alignment}
Optical flow estimation is an analogous problem to 3D volume registration for 2D images. Optical flow algorithms return a dense displacement vector field depicting small displacements between a 2D image pair. Traditional optical flow approaches typically solve an optimization problem similar to~\eqref{eqn:energy} using variational methods~\cite{brox2004,horn1980,sun2010}. Extensions that better handle large displacements or dramatic changes in appearance include feature-based matching~\cite{brox2011,liu2011} and nearest neighbor fields~\cite{chen2013}. 

Several learning-based approaches to dense 2D image alignment have been proposed. One study learns a low-dimensional basis for optical flow in natural images using PCA~\cite{wulff2015}. Other recent studies in optical flow learn a parametric function using convolutional neural networks~\cite{dosovitskiy2015,weinzaepfel2013}. Unfortunately, these methods require ground truth registrations during training. The spatial transform layer enables neural networks to perform global parametric 2D image alignment without requiring supervised labels~\cite{jaderberg2015}. 
The layer has since been used for dense spatial transformations as well~\cite{park2017,zhou2016}. We extend the spatial transformer to the 3D setting in our work.
\begin{figure*}[h!]
\begin{center}
\includegraphics[width=\textwidth]{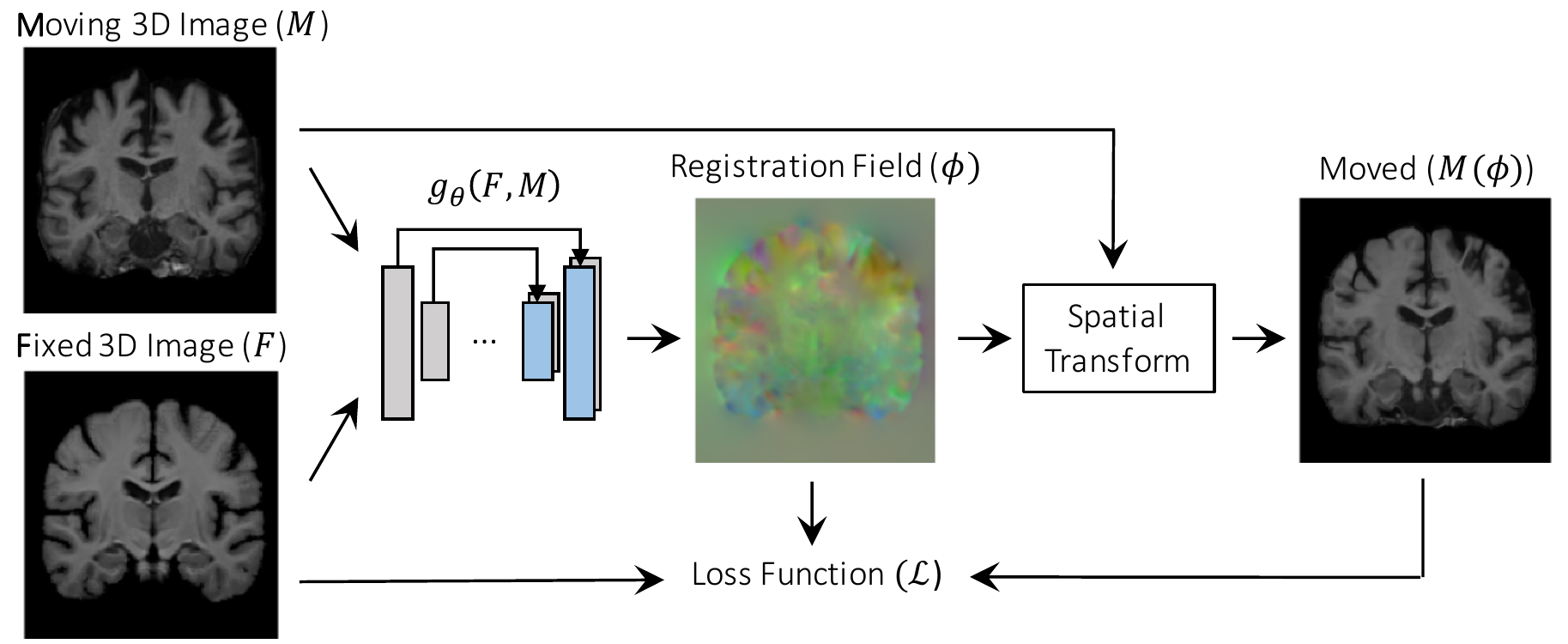}
\end{center}
\caption{Overview of our method. We learn parameters for a function $g$ that registers one 3D volume ($M$) to a second, fixed volume ($F$). During training, we warp $M$ with $\phi$ using a spatial transformer function. Our loss compares $M_{\phi}$ and $F$ and enforces smoothness of $\phi$.}
\label{fig:overview}
\end{figure*}

\section{Method}
Let $F,M$ be two image volumes defined over a $n$-D spatial domain $\Omega \subset \mathcal{R}^n$. For the rest of this paper, we focus on the case $n=3$. For simplicity we assume that $F$ and $M$ contain single-channel, grayscale data. We also assume that $F$ and $M$ are affinely aligned as a preprocessing step, so that the only source of misalignment between the volumes is nonlinear. Many packages are available for rapid affine alignment.

We model a function $g_{\theta}(F,M) = \phi$ using a convolutional neural network (CNN), where $\phi$ is a registration field and $\theta$ are learnable parameters of $g$. For each voxel $p \in \Omega$, $\phi (p)$ is a location such that $F(p)$ and $M(\phi (p))$ define similar anatomical locations.

Fig.~\ref{fig:overview} presents an overview of our method. Our network takes $M$ and $F$ as input, and computes $\phi$ based on a set of parameters $\theta$, the kernels of the convolutional layers. We warp $M(p)$ to $M(\phi (p))$ using a spatial transformation function, enabling the model to evaluate the similarity of $M(\phi)$ and $F$ and update $\theta$. 

We use stochastic gradient descent to find optimal parameters $\hat{\theta}$ by minimizing an expected loss function $\mathcal{L}(\cdot,\cdot,\cdot)$, similar to~\eqref{eqn:loss}, using a training dataset:
\begin{align}
\hat{\theta} &= \argmin_{\theta} \left[\mathbb{E}_{(F,M)\sim D}\left[\mathcal{L}\left(F,M,g_{\theta}(F,M)\right)\right]\right], 
\label{eqn:poploss}
\end{align}

\noindent where $D$ is the dataset distribution. We learn $\hat{\theta}$ by aligning volume pairs sampled from $D$. Importantly, we do not require supervised information such as ground truth registration fields or anatomical landmarks. Given an unseen $M$ and $F$ during test time, we obtain a registration field by evaluating $g$. We describe our model, which we call \emph{\algname}, in the next sections. 

\subsection{\algname{}  CNN Architecture}
The parametrization of $g$ is based on a convolutional neural network architecture similar to UNet~\cite{isola2017,ronneberger2015}. The network consists of an encoder-decoder with skip connections that is responsible for generating $\phi$ given $M$ and $F$.


\begin{figure}[h!]
\begin{center}
\includegraphics[width=\linewidth]{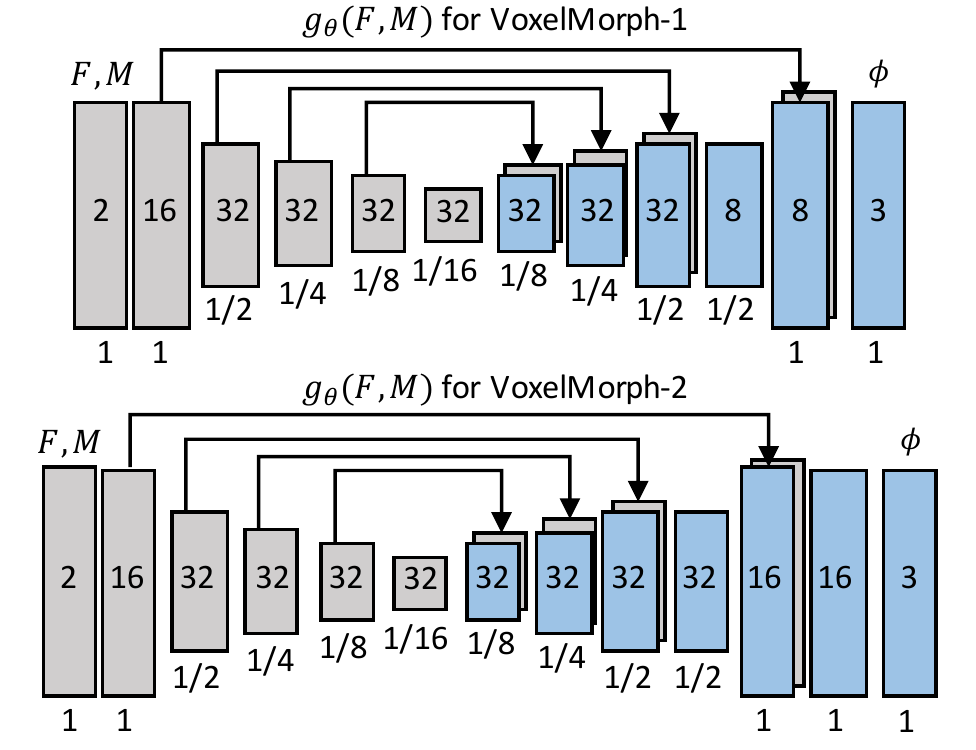}
\end{center}
\caption{Proposed convolutional architectures implementing $g_{\theta}(F,M)$. Each rectangle represents a 3D volume. The number of channels is shown inside the rectangle, and the spatial resolution with respect to the input volume is printed underneath. \algname-2 uses a larger architecture, using one extra convolutional layer at the output resolution, and more channels for later layers.}
\label{fig:unets}
\end{figure}

Fig.~\ref{fig:unets} depicts two variants of the proposed architectures that tradeoff between registration accuracy and computation time. Both take a single input formed by concatenating $M$ and $F$ into a 2-channel 3D image. In our experiments, the input is of size $160\times192\times224\times2$. We apply 3D convolutions followed by Leaky ReLU activations in both the encoder and decoder stages, with a convolutional kernel size of $3\times3\times3$. The convolutional layers capture hierarchical features of the input image pair necessary to estimate the correspondence $\phi$. In the encoder, we use strided convolutions to reduce the spatial dimensions in half until the smallest layer is reached. Successive layers of the encoder operate over coarser representations of the input, similar to the image pyramid used in traditional image registration work. 

The receptive fields of the convolutional kernels of the smallest layer should be at least as large as the maximum expected displacement between corresponding voxels in $M$ and $F$. The smallest layer applies convolutions over a volume $(1/16)^3$ the size of the input images. In the decoding stage, we alternate between upsampling, convolutions (followed by Leaky ReLU activations) and concatenating skip connections. Skip connections propagate features learned during the encoding stages directly to layers generating the registration. The output of the decoder, $\phi$, is of size $160\times192\times224\times3$ in our experiments.

Successive layers of the decoder operate on finer spatial scales, enabling precise anatomical alignment. However, these convolutions are applied to the largest image volumes, which is computationally expensive. We explore this tradeoff using two architectures, \algname-1 and \algname-2, that differ in size at the end of the decoder (see Fig.~\ref{fig:unets}). \algname-1 uses one less layer at the final resolution and fewer channels over its last three layers.

\subsection{Spatial Transformation Function}
The proposed method learns optimal parameter values in part by minimizing differences between $M(\phi)$ and $F$. In order to use standard gradient-based methods, we construct a differentiable operation based on spatial transformer networks to compute $M(\phi)$~\cite{jaderberg2015}. 

For each voxel $p$, we compute a (subpixel) voxel location $\phi(p)$ in $M$. Because image values are only defined at integer locations, we linearly interpolate the values at the eight neighboring voxels. That is, we perform:

\small
\begin{equation}
M(\phi(p)) =\hspace{-0.25cm}\sum_{q \in \mathcal{Z}(\phi (p))}\hspace{-0.25cm}M(q)\hspace{-0.25cm}\prod_{d \in \{x,y,z\}}\hspace{-0.25cm}(1-|\phi_d(p)-q_d|),\\
\end{equation}
\normalsize
\noindent where $\mathcal{Z}(\phi(p))$ are the voxel neighbors of $\phi (p)$. Because the operations are differentiable almost everywhere, we can backpropagate errors during optimization.  

\subsection{Loss Function}
The proposed method works with any differentiable loss. In this section, we formulate an example of a popular loss function $\mathcal{L}$ of the form~\eqref{eqn:loss}, consisting of two components: $\mathcal{L}_{sim}$ that penalizes differences in appearance, and $\mathcal{L}_{smooth}$ that penalizes local spatial variations in $\phi$. In our experiments, we set $\mathcal{L}_{sim}$ to the negative local cross-correlation of $M(\phi)$ and $F$, a popular metric that is robust to intensity variations often found across scans and datasets.

Let $\hat{F}(p)$ and $\hat{M}(\phi (p))$ denote images with local mean intensities subtracted out. We compute local means over a $n^3$ volume, with $n=9$ in our experiments. We write the local cross-correlation of $F$ and $M(\phi)$, as:

\footnotesize
\begin{align}
&CC(F,M(\phi))= \nonumber\\
&\sum\limits_{p \in \Omega} \frac{\left(\sum\limits_{p_i} (F(p_i) - \hat{F}(p))(M(\phi(p_i)) - \hat{M}(\phi(p)))\right)^2}
{\left(\sum\limits_{p_i} (F(p_i) - \hat{F}(p)) \right)\left(\sum\limits_{p_i} (M(\phi(p_i)) - \hat{M}(\phi(p)))\right)}, 
\end{align}
\normalsize

\noindent where $p_i$ iterates over a $n^3$ volume around $p$. A higher CC indicates a better alignment, yielding the loss function: $\mathcal{L}_{sim}(F,M,\phi) = - CC(F,M(\phi))$. We compute CC efficiently using only convolutional operations over $M(\phi)$ and~$F$.

Minimizing $\mathcal{L}_{sim}$ will encourage $M(\phi)$ to approximate~$F$, but may generate a discontinuous $\phi$. We encourage a smooth $\phi$ using a diffusion regularizer on its spatial gradients:

\begin{align}
\mathcal{L}_{smooth}(\phi) = \sum_{p \in \Omega} \lVert \nabla \phi(p) \rVert ^2.
\end{align}

\noindent We approximate spatial gradients using differences between neighboring voxels. The complete loss is therefore:

\begin{align}
\mathcal{L}(F,M,\phi) =-CC(F,M(\phi)) + \lambda \sum_{p \in \Omega} \lVert \nabla \phi(p) \rVert ^2,
\end{align}

\noindent where $\lambda$ is a regularization parameter.
\section{Experiments}

\subsection{Dataset}
We demonstrate our method on the task of brain MRI registration. We use a large-scale, multi-site, multi-study dataset of 7829 T1‐weighted brain MRI scans from eight publicly available datasets: ADNI~\cite{mueller2005ways}, OASIS~\cite{marcus2007open}, ABIDE~\cite{di2014autism}, ADHD200~\cite{milham2012adhd}, MCIC~\cite{gollub2013mcic}, PPMI~\cite{marek2011parkinson}, HABS~\cite{dagley2015harvard}, and Harvard GSP~\cite{holmes2015brain}. Acquisition details, subject age ranges and health conditions are different for each dataset. All scans were resampled to a $256\times 256 \times 256$ grid with $1$mm isotropic voxels. We carry out standard pre‐processing steps, including affine spatial normalization and brain extraction for each scan using FreeSurfer~\cite{fischl2012}, and crop the resulting images to $160\times 192 \times 224$. All MRIs were also anatomically segmented with FreeSurfer, and we applied quality control (QC) using visual inspection to catch gross errors in segmentation results. We use the resulting segmentation maps in evaluating our registration as described below. We split our dataset into 7329, 250, and 250 volumes for train, validation, and test sets respectively, although we highlight that we do not use any supervised information at any stage.

\begin{figure}[h!]
\begin{center}
\includegraphics[width=\linewidth]{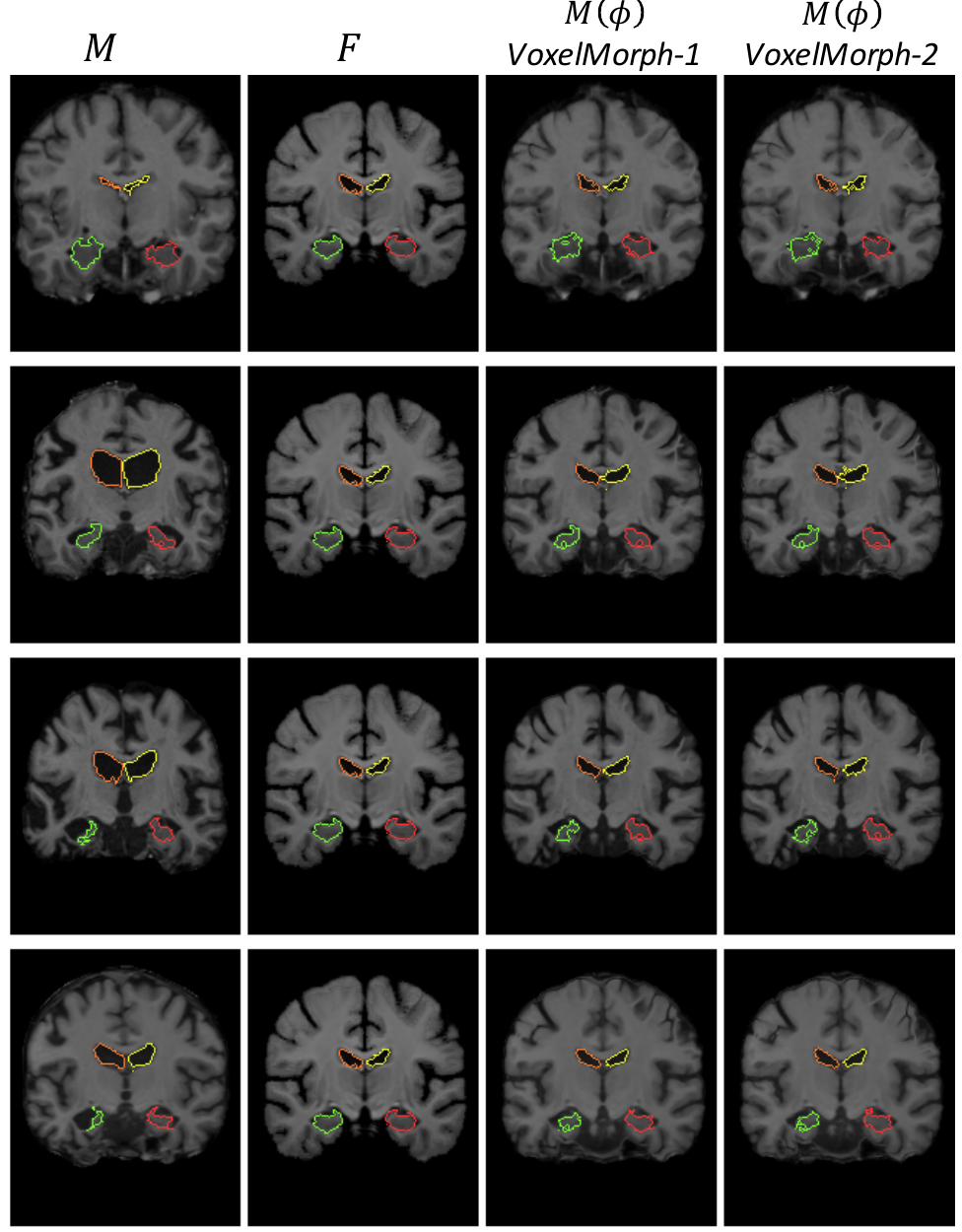}
\end{center}
\caption{Example MR coronal slices extracted from input pairs (columns 1-2), and resulting $M(\phi)$ for \algname-1 and \algname-2, with overlaid boundaries of the ventricles (yellow, orange) and hippocampi (red, green). A good registration will cause structures in $M(\phi)$ to look similar to structures in $F$. Our networks handle large changes in shapes, such as the ventricles in row 2 and the left hippocampi in rows 3-4.}
\label{fig:reg_examples}
\end{figure}

We focus on atlas-based registration, in which we compute a registration field between an atlas, or reference volume, and each volume in our dataset.  Atlas-based registration is a common formulation in population analysis, where inter-subject registration is a core problem. The atlas represents a reference, or average volume, and is usually constructed by jointly and repeatedly aligning a dataset of brain MR volumes and averaging them together. We use an atlas computed using an external dataset~\cite{fischl2012,sridharan2013}. Each input volume pair consists of the atlas (image $F$) and a random volume from the dataset (image $M$). Columns 1-2 of Fig.~\ref{fig:reg_examples} show example image pairs from the dataset using the same fixed atlas for all examples. All figures that depict brains in this paper show 2D coronal slices for visualization purposes only. All registration is done in 3D.

\subsection{Dice Score}
Obtaining dense \emph{ground truth} registration for these data is not well-defined since many registration fields can yield similar looking warped images. We evaluate our method using volume overlap of anatomical segmentations. We include any anatomical structures that are at least $100$ voxels in volume for all test subjects, resulting in 29 structures. If a registration field $\phi$ represents accurate anatomical correspondences, we expect the regions in $F$ and $M(\phi)$ corresponding to the same anatomical structure to overlap well (see Fig.~\ref{fig:reg_examples} for examples). Let $S^k_{F},S^k_{M(\phi)}$ be the set of voxels of structure $k$ for $F$ and $M(\phi)$, respectively. We measure the accuracy of our method using the Dice score~\cite{dice1945}, which quantifies the volume overlap between two structures:

\begin{equation}
\text{Dice}(S^k_{M(\phi)}, S^k_F) = 2*\frac{S^k_{M(\phi)} \cap S^k_F}{| S^k_{M(\phi)}| + |S^k_F|}.
\end{equation}
\noindent A Dice score of $1$ indicates that the structures are identical, and a score of 0 indicates that there is no overlap.

\subsection{Baseline Methods}
We compare our approach to Symmetric Normalization (SyN)~\cite{avants2008}, the top-performing registration algorithm in a comparative study~\cite{klein2009}. We use the SyN implementation in the publicly available ANTs software package~\cite{avants2011}, with a cross-correlation similarity measure. Throughout our work with medical images, we found the default ANTs smoothness parameters to be sub-optimal for our purposes. We obtained improved parameters using a wide parameter sweep across a multiple of datasets, and use those in these experiments.

\subsection{Implementation}
We implement our networks using Keras~\cite{chollet2015} with a Tensorflow backend~\cite{abadi2016}. We use the ADAM optimizer~\cite{kingma2014} with a learning rate of $1e^{-4}$. To reduce memory usage, each training batch consists of one pair of volumes. We train separate networks with different $\lambda$ values until convergence. We select the network that optimizes Dice score on our validation set, and report results on our held-out test set. Our code and model parameters are available online at https://github.com/balakg/voxelmorph.

\subsection{Results}

\subsubsection{Accuracy}
\begin{figure*}[h!]
\begin{center}
\includegraphics[width=\linewidth]{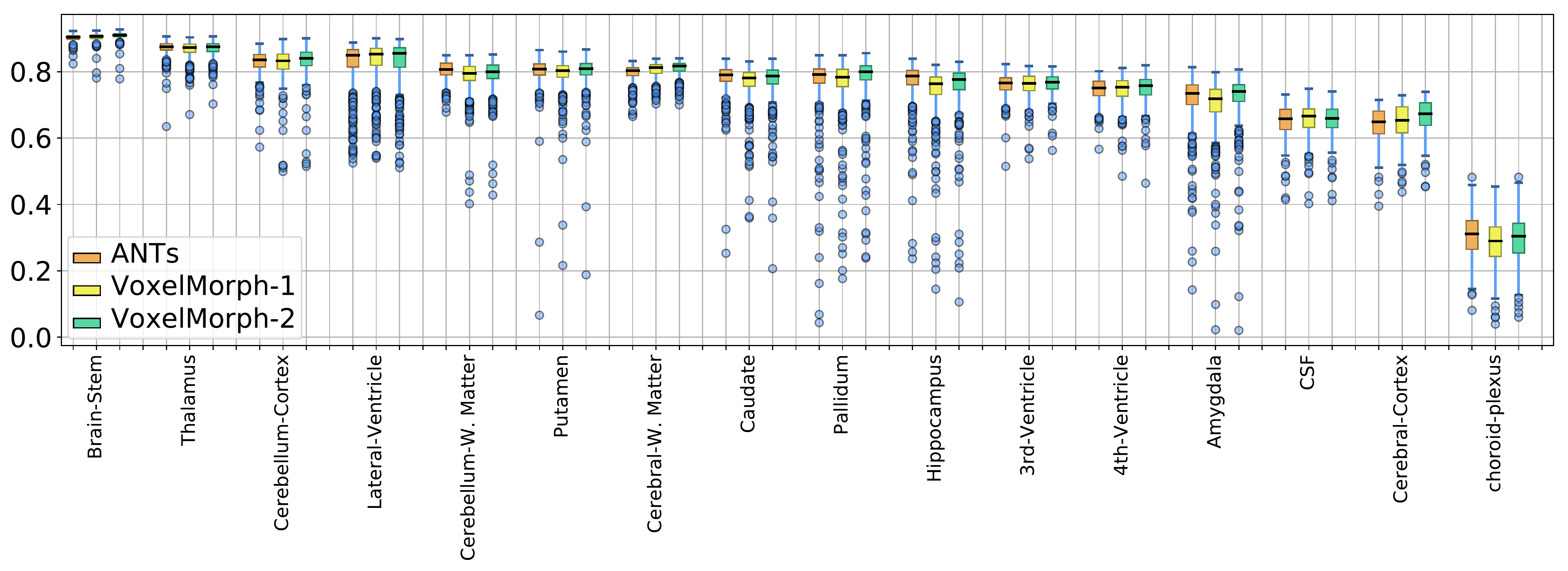}
\end{center}
\caption{Boxplots of Dice scores for anatomical structures for \algname-1, \algname-2 and ANTs. We combine structures with separate left and right brain hemispheres into one structure for this visualization. Structures are ordered by average ANTs Dice score.}
\label{fig:boxplot}
\end{figure*}

Table~\ref{tbl:results} shows average Dice scores over all subjects and structures for ANTs, the proposed \algname{} architectures, and a baseline of only global affine alignment. \algname{} models perform comparably to ANTs, and \algname-2 performs slightly better than \algname-1. All three improve significantly on affine alignment. We visualize the distribution of Dice scores for each structure as boxplots in Fig.~\ref{fig:boxplot}. For visualization purposes, we combine same structures from the two hemispheres, such as the left and right white matter. The \algname{} models achieve comparable Dice measures to ANTs for all structures, performing slightly better than ANTs on some structures such as cerebral white matter, and worse on others such as the hippocampi.

\begin{table}[t!]
\small
\centering
\caption{Average Dice scores and runtime results for affine alignment, ANTs, \algname-1, \algname-2. Standard deviations are in parentheses. The average Dice score is computed over all structures and subjects. Timing is computed after preprocessing. Our networks yield comparable results to ANTs in Dice score, while operating orders of magnitude faster during testing. To our knowledge, ANTs does not have a GPU implementation.}
\begin{tabular}{c c c c}
Method&Avg. Dice&GPU sec&CPU sec\\
\hline
Affine only&0.567 (0.157)&0&0\\
ANTs&0.749 (0.135)&-&9059 (2023)\\
\algname-1&0.742 (0.139)&0.365 (0.012)&57(1)\\
\algname-2&0.750 (0.137)&0.554 (0.017)&144 (1)\\
\end{tabular}
\label{tbl:results}
\end{table}   
\normalsize

\subsubsection{Runtime}
Table~\ref{tbl:results} presents runtime results using an Intel Xeon (E5-2680) CPU, and a NVIDIA TitanX GPU. We report the elapsed time for computations following the affine alignment preprocessing step, which all of the presented methods share, and requires just a few minutes on a CPU. ANTs requires roughly two or more hours of CPU time. \algname-1 and \algname-2 are 60+ and 150+ times faster on average using the CPU. ANTs runtimes vary widely, because its convergence depends on the difficulty of the alignment task. When using the GPU, our networks compute a registration in under a second. To our knowledge, there is no publicly available ANTs implementation for GPUs.

\subsubsection{Training and Testing on a Sub-population}
The results in the previous sections combine multiple datasets consisting of different population types, resulting in a trained model that generalizes well to a range of subjects. In this section, we model parameters specific to a subpopulation, demonstrating the ability of tailoring our approach to particular tasks. We train using ABIDE subject scans, and evaluate test performance on unseen ABIDE scans. ABIDE contains scans of subjects with autism and controls, and includes a wide age range, with a median age of 15 years. In Table~\ref{tbl:results2} we compare the results to those of the models trained on all datasets, presented in the previous section. The dataset-specific networks achieve a $1.5\%$ Dice score improvement. 

\small
\begin{table}[b!]
\centering
\caption{Average Dice scores on ABIDE scans, when trained on all datasets (column 2) and ABIDE scans only (column 3). We achieve roughly $1.5\%$ better scores when training on ABIDE only.}
\begin{tabular}{c c c}
&Avg. Dice&Avg. Dice\\
Method&(Train on All)&(Train on ABIDE)\\
\hline
\algname-1&0.715(0.140)&0.729(0.142)\\
\algname-2&0.718(0.141)&0.734(0.140)\\
\end{tabular}
\label{tbl:results2}
\end{table}   
\normalsize

\subsubsection{Regularization Analysis}
\begin{figure}
    \centering
    \begin{subfigure}[b]{\linewidth}
        \includegraphics[width=\linewidth]{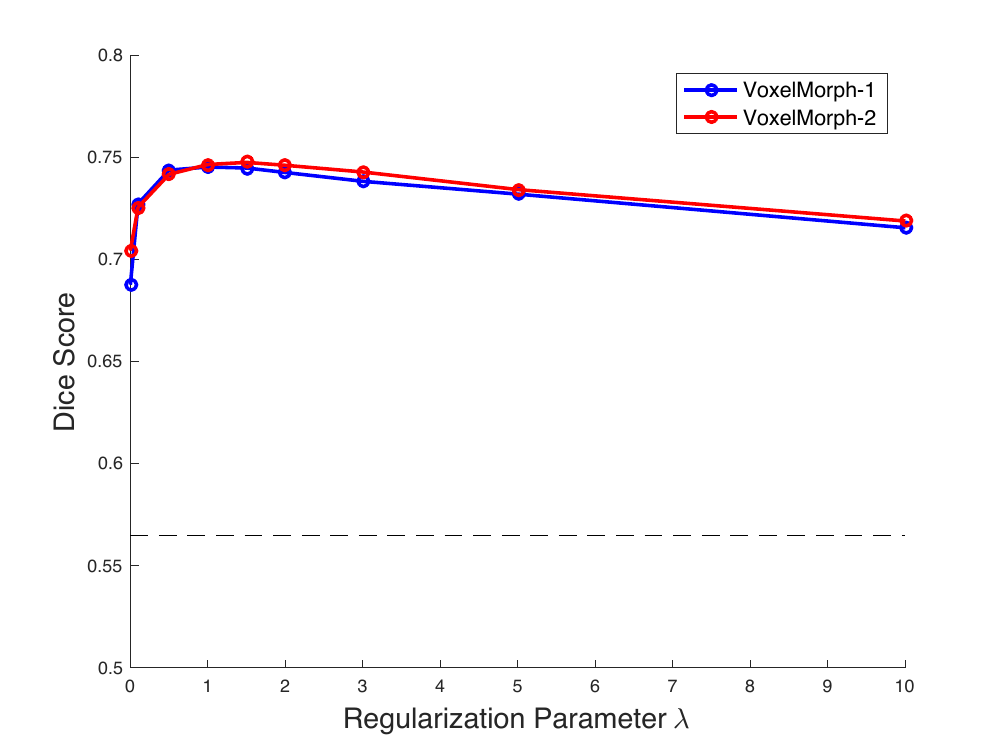}
        \caption{}
        \label{fig:auc}
    \end{subfigure}
\\
    \begin{subfigure}[b]{\linewidth}
        \includegraphics[width=\linewidth]{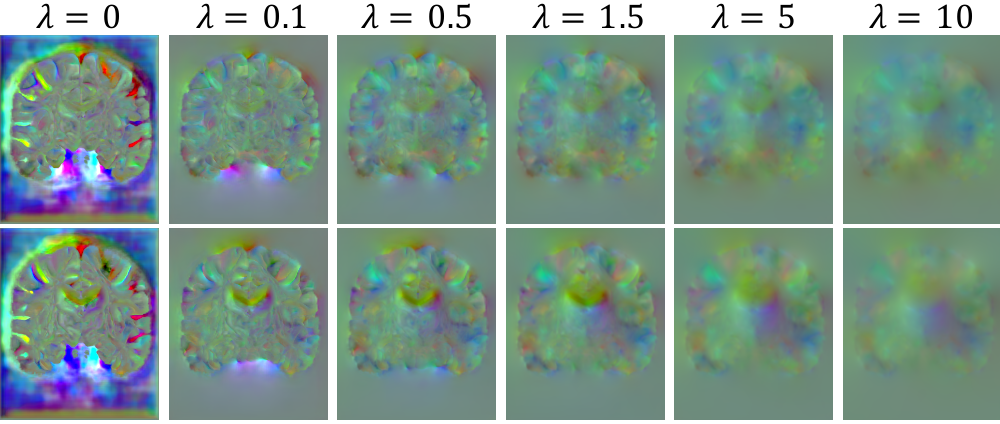}
        \caption{}
        \label{fig:flow}
    \end{subfigure}
    
     \caption{(a) Effect of varying the regularization parameter $\lambda$ on Dice score. The best results occur when $\lambda=1$ for \algname-1 and $\lambda=1.5$ for \algname-2. Also shown are Dice scores when applying only affine registration. (b) Examples of \algname-2 registration fields for a 2D coronal slice, for different values of $\lambda$. Each row is a different scan. We clip the $x,y,z$ displacements to $[-10,10]$, rescale them to $[0,1]$, and place them in RGB channels. As $\lambda$ increases, the registration field becomes smoother across structural boundaries.}
\end{figure}

Fig.~\ref{fig:auc} presents average Dice scores for the validation set for different values of the smoothing parameter $\lambda$. As a baseline, we display Dice score of the affinely aligned scans. The optimal Dice scores occur when $\lambda=1$ for \algname-1 and $\lambda=1.5$ for \algname-2. However, the results vary slowly over a large range of $\lambda$ values, showing that our model is robust to choice of $\lambda$. Interestingly, even setting $\lambda = 0$, which enforces no regularization, results in a significant improvement over affine registration. This is likely due to the fact that the optimal network parameters $\hat{\theta}$ need to register all pairs in the training set well, giving an implicit regularization. Fig.~\ref{fig:flow} shows example registration fields at a coronal slice with different regularization values. For low $\lambda$, the field can change dramatically across edges and structural boundaries.

\section{Discussion}
Our model is able to perform on par with the state-of-the-art ANTs registration package while requiring far less computation time to register test volume pairs. While our method learns general features about the data necessary for registration, it can adapt these parameters to specific subpopulations. When training on the ABIDE dataset only, we obtain improved Dice scores on test ABIDE scans compared to training on a dataset from several sources exhibiting different health conditions and variations in acquisition. This result shows that some of our model's parameters are learning properties specific to the training images.

We present two models which trade off in accuracy and computation time. The smaller architecture, \algname{}-1, runs significantly faster on the CPU and is less than $1$ Dice point worse than \algname{}-2. This enables an application-specific decision. An advantage of our model is that it is easy to explore this tradeoff by changing the number of convolutional layers and channels of the network, which can be considered as hyperparameters. We selected these hyperparameters by experimenting on training and validation data, and they could be adapted to other tasks.  

We quantify accuracy in this study using Dice score, which acts as a proxy measure of registration accuracy. While our models achieve comparable Dice scores, ANTs produces diffeomorphic registrations, which are not guaranteed by our models. Diffeomorphic fields have attractive properties like invertibility and topology-preservation that are useful in some analyses. This presents an exciting area of future work for learning-based registration.

Our method replaces a costly optimization problem for each test image pair, with one function optimization aggregated over a dataset during a training phase. This idea is applicable to a wide variety of problems traditionally relying on complex, non-learning-based optimization algorithms for each input. Our network implementations needed a one-time training period of a few days on a single NVIDIA TITANX GPU, but less than a second to register a test pair of images. Given the growing availability of image data, our solution is preferable to a non-learning-based approach, and sorely-needed to facilitate fast medical image analyses. 
\section{Conclusion}
This paper presents an unsupervised learning-based approach to medical image registration, that requires no supervised information such as ground truth registration fields or anatomical landmarks. The approach obtains similar registration accuracy to state-of-the-art 3D image registration on a large-scale, multi-study MR brain dataset, while operating orders of magnitude faster. Model analysis shows that our model is robust to regularization parameter, can be tailored to different data populations, and can be easily modified to explore accuracy and runtime tradeoffs. Our method promises to significantly speed up medical image analysis and processing pipelines, while facilitating novel directions in learning-based registration.

{\small
\bibliographystyle{ieee}
\bibliography{egbib}

\begin{thebibliography}{10}\itemsep=-1pt

\bibitem{abadi2016}
M.~Abadi et~al.
\newblock Tensorflow: Large-scale machine learning on heterogeneous distributed
  systems.
\newblock {\em arXiv preprint arXiv:1603.04467}, 2016.

\bibitem{ashburner2007}
J.~Ashburner.
\newblock A fast diffeomorphic image registration algorithm.
\newblock {\em Neuroimage}, 38(1):95--113, 2007.

\bibitem{ashburner2000}
J.~Ashburner and K.~Friston.
\newblock Voxel-based morphometry-the methods.
\newblock {\em Neuroimage}, 11:805--821, 2000.

\bibitem{avants2008}
B.~B. Avants et~al.
\newblock Symmetric diffeomorphic image registration with cross-correlation:
  evaluating automated labeling of elderly and neurodegenerative brain.
\newblock {\em Medical image analysis}, 12(1):26--41, 2008.

\bibitem{avants2011}
B.~B. Avants et~al.
\newblock A reproducible evaluation of ants similarity metric performance in
  brain image registration.
\newblock {\em Neuroimage}, 54(3):2033--2044, 2011.

\bibitem{bajcsy1989}
R.~Bajcsy and S.~Kovacic.
\newblock Multiresolution elastic matching.
\newblock {\em Computer Vision, Graphics, and Image Processing}, 46:1--21,
  1989.

\bibitem{beg2005}
M.~F. Beg et~al.
\newblock Computing large deformation metric mappings via geodesic flows of
  diffeomorphisms.
\newblock {\em Int. J. Comput. Vision}, 61:139--157, 2005.

\bibitem{brox2004}
T.~Brox et~al.
\newblock High accuracy optical flow estimation based on a theory for warping.
\newblock {\em European Conference on Computer Vision (ECCV)}, pages 25--36,
  2004.

\bibitem{brox2011}
T.~Brox and J.~Malik.
\newblock Large displacement optical flow: Descriptor matching in variational
  motion estimation.
\newblock {\em IEEE Trans. Pattern Anal. Mach. Intell.}, 33(3):500--513, 2011.

\bibitem{chen2013}
Z.~Chen et~al.
\newblock Large displacement optical flow from nearest neighbor fields.
\newblock {\em IEEE Conference on Computer Vision and Pattern Recognition
  (CVPR)}, pages 2443--2450, 2013.

\bibitem{chollet2015}
F.~Chollet et~al.
\newblock Keras.
\newblock \url{https://github.com/fchollet/keras}, 2015.

\bibitem{dagley2015harvard}
A.~Dagley et~al.
\newblock Harvard aging brain study: dataset and accessibility.
\newblock {\em NeuroImage}, 2015.

\bibitem{dalca2016}
A.~V. Dalca et~al.
\newblock Patch-based discrete registration of clinical brain images.
\newblock In {\em International Workshop on Patch-based Techniques in Medical
  Imaging}, pages 60--67. Springer, 2016.

\bibitem{devos2017}
B.~de~Vos et~al.
\newblock End-to-end unsupervised deformable image registration with a
  convolutional neural network.
\newblock In {\em Deep Learning in Medical Image Analysis and Multimodal
  Learning for Clinical Decision Support}, pages 204--212. 2017.

\bibitem{dice1945}
L.~R. Dice.
\newblock Measures of the amount of ecologic association between species.
\newblock {\em Ecology}, 26(3):297--302, 1945.

\bibitem{dosovitskiy2015}
A.~Dosovitskiy et~al.
\newblock Flownet: Learning optical flow with convolutional networks.
\newblock In {\em IEEE International Conference on Computer Vision (ICCV)},
  pages 2758--2766, 2015.

\bibitem{fischl2012}
B.~Fischl.
\newblock Freesurfer.
\newblock {\em Neuroimage}, 62(2):774--781, 2012.

\bibitem{glocker2008}
B.~Glocker et~al.
\newblock Dense image registration through mrfs and efficient linear
  programming.
\newblock {\em Medical image analysis}, 12(6):731--741, 2008.

\bibitem{gollub2013mcic}
R.~L. Gollub et~al.
\newblock The mcic collection: a shared repository of multi-modal, multi-site
  brain image data from a clinical investigation of schizophrenia.
\newblock {\em Neuroinformatics}, 11(3):367--388, 2013.

\bibitem{holmes2015brain}
A.~J. Holmes et~al.
\newblock Brain genomics superstruct project initial data release with
  structural, functional, and behavioral measures.
\newblock {\em Scientific data}, 2, 2015.

\bibitem{horn1980}
B.~K. Horn and B.~G. Schunck.
\newblock Determining optical flow.
\newblock 1980.

\bibitem{isola2017}
P.~Isola et~al.
\newblock Image-to-image translation with conditional adversarial networks.
\newblock {\em arXiv preprint}, 2017.

\bibitem{jaderberg2015}
M.~Jaderberg et~al.
\newblock Spatial transformer networks.
\newblock In {\em Advances in neural information processing systems}, pages
  2017--2025, 2015.

\bibitem{kingma2014}
D.~P. Kingma and J.~Ba.
\newblock {ADAM}: A method for stochastic optimization.
\newblock {\em arXiv preprint arXiv:1412.6980}, 2014.

\bibitem{klein2009}
A.~Klein et~al.
\newblock Evaluation of 14 nonlinear deformation algorithms applied to human
  brain mri registration.
\newblock {\em Neuroimage}, 46(3):786--802, 2009.

\bibitem{krebs2017}
J.~Krebs et~al.
\newblock Robust non-rigid registration through agent-based action learning.
\newblock In {\em International Conference on Medical Image Computing and
  Computer-Assisted Intervention (MICCAI)}, pages 344--352. Springer, 2017.

\bibitem{li2017}
H.~Li and Y.~Fan.
\newblock Non-rigid image registration using fully convolutional networks with
  deep self-supervision.
\newblock {\em arXiv preprint arXiv:1709.00799}, 2017.

\bibitem{liu2011}
C.~Liu et~al.
\newblock {SIFT} flow: Dense correspondence across scenes and its applications.
\newblock {\em IEEE Trans. Pattern Anal. Mach. Intell.}, 33(5):978--994, 2011.

\bibitem{marcus2007open}
D.~S. Marcus et~al.
\newblock Open access series of imaging studies (oasis): cross-sectional mri
  data in young, middle aged, nondemented, and demented older adults.
\newblock {\em Journal of cognitive neuroscience}, 19(9):1498--1507, 2007.

\bibitem{marek2011parkinson}
K.~Marek et~al.
\newblock The parkinson progression marker initiative.
\newblock {\em Progress in neurobiology}, 95(4):629--635, 2011.

\bibitem{di2014autism}
A.~D. Martino et~al.
\newblock The autism brain imaging data exchange: towards a large-scale
  evaluation of the intrinsic brain architecture in autism.
\newblock {\em Molecular psychiatry}, 19(6):659--667, 2014.

\bibitem{milham2012adhd}
M.~P. Milham et~al.
\newblock The {ADHD-200} consortium: a model to advance the translational
  potential of neuroimaging in clinical neuroscience.
\newblock {\em Frontiers in systems neuroscience}, 6:62, 2012.

\bibitem{mueller2005ways}
S.~G. Mueller et~al.
\newblock Ways toward an early diagnosis in alzheimer's disease: the
  alzheimer's disease neuroimaging initiative (adni).
\newblock {\em Alzheimer's \& Dementia}, 1(1):55--66, 2005.

\bibitem{park2017}
E.~Park et~al.
\newblock Transformation-grounded image generation network for novel {3D} view
  synthesis.
\newblock In {\em IEEE Conference on Computer Vision and Pattern Recognition
  (CVPR)}, pages 702--711, 2017.

\bibitem{rohe2017}
M.-M. Roh{\'e} et~al.
\newblock Svf-net: Learning deformable image registration using shape matching.
\newblock In {\em International Conference on Medical Image Computing and
  Computer-Assisted Intervention (MICCAI)}, pages 266--274. Springer, 2017.

\bibitem{ronneberger2015}
O.~Ronneberger et~al.
\newblock U-net: Convolutional networks for biomedical image segmentation.
\newblock In {\em International Conference on Medical Image Computing and
  Computer-Assisted Intervention (MICCAI)}, pages 234--241. Springer, 2015.

\bibitem{rueckert1999}
D.~Rueckert et~al.
\newblock Nonrigid registration using free-form deformation: Application to
  breast mr images.
\newblock {\em IEEE Transactions on Medical Imaging}, 18(8):712--721, 1999.

\bibitem{shen2002}
D.~Shen and C.~Davatzikos.
\newblock Hammer: Hierarchical attribute matching mechanism for elastic
  registration.
\newblock {\em IEEE Transactions on Medical Imaging}, 21(11):1421--1439, 2002.

\bibitem{sokooti2017}
H.~Sokooti et~al.
\newblock Nonrigid image registration using multi-scale 3d convolutional neural
  networks.
\newblock In {\em International Conference on Medical Image Computing and
  Computer-Assisted Intervention (MICCAI)}, pages 232--239. Springer, 2017.

\bibitem{sridharan2013}
R.~Sridharan et~al.
\newblock Quantification and analysis of large multimodal clinical image
  studies: Application to stroke.
\newblock In {\em International Workshop on Multimodal Brain Image Analysis},
  pages 18--30. Springer, 2013.

\bibitem{sun2010}
D.~Sun et~al.
\newblock Secrets of optical flow estimation and their principles.
\newblock {\em IEEE Conference on Computer Vision and Pattern Recognition
  (CVPR)}, pages 2432--2439, 2010.

\bibitem{thirion1998}
J.~Thirion.
\newblock Image matching as a diffusion process: an analogy with maxwell's
  demons.
\newblock {\em Medical Image Analysis}, 2(3):243--260, 1998.

\bibitem{weinzaepfel2013}
P.~Weinzaepfel et~al.
\newblock Deepflow: Large displacement optical flow with deep matching.
\newblock In {\em IEEE International Conference on Computer Vision (ICCV)},
  pages 1385--1392, 2013.

\bibitem{wulff2015}
J.~Wulff and M.~J. Black.
\newblock Efficient sparse-to-dense optical flow estimation using a learned
  basis and layers.
\newblock In {\em IEEE Conference on Computer Vision and Pattern Recognition
  (CVPR)}, pages 120--130, 2015.

\bibitem{yang2017}
X.~Yang et~al.
\newblock Quicksilver: Fast predictive image registration--a deep learning
  approach.
\newblock {\em NeuroImage}, 158:378--396, 2017.

\bibitem{zhou2016}
T.~Zhou et~al.
\newblock View synthesis by appearance flow.
\newblock {\em European Conference on Computer Vision (ECCV)}, pages 286--301,
  2016.

\end{thebibliography}
}

\end{document}